# English to Bangla Machine Translation Using Recurrent Neural Network

Shaykh Siddique, Tahmid Ahmed, Md. Rifayet Azam Talukder, and Md. Mohsin Uddin

*Abstract*—The applications of recurrent neural networks in machine translation are increasing in natural language processing. Besides other languages, Bangla language contains a large amount of vocabulary. Improvement of English to Bangla machine translation would be a significant contribution to Bangla Language processing. This paper describes an architecture of English to Bangla machine translation system. The system has been implemented with the encoder-decoder recurrent neural network. The model uses a knowledge-based context vector for the mapping of English and Bangla words. Performances of the model based on activation functions are measured here. The best performance is achieved for the linear activation function in encoder layer and the tanh activation function in decoder layer. From the execution of GRU and LSTM layer, GRU performed better than LSTM. The attention layers are enacted with softmax and sigmoid activation function. The approach of the model outperforms the previous state-of-the-art systems in terms of cross-entropy loss metrics. The reader can easily find out the structure of the machine translation of English to Bangla and the efficient activation functions from the paper.

## I. INTRODUCTION

### A. Background

Translation means converting one language to another language and an automated translation system plays an important role as a translator. A translation could be a word by word translation and another one is translation by a sentence. In a sentence translation, more information was gotten rather than word by word translation. In this paper, it represents the main focus which is to translate from English to Bangla by using a machine learning algorithm. Machine translation refers to the translation of text or speech. The main task or aim is to translate from the English language to Bangla language with approximately 228 million native speakers and another 37 million as second language speakers. Bangla is the fifth most-spoken native language and the seventh most spoken language by the total number of speakers in the world [1]. Natural language processing is used to make the machine intelligent. The way of language processing is enriching day by day. Many studies defined the architecture for natural language processing [2], but some deal with the improvement of English to Bangla language translation. A team works in Tense Based Structure of English to Bangla translation [3]. Another study is on simple sentence structure and comparison of different machine translation systems [4]. Still, now there is a lack of studies with complex sentence structure and recurrent meaning of a sentence.

Translating using the machine is important because as new data will add to the model it will be able to adopt the changes independently. Moreover, a machine can handle multi-dimensional data as well as multi-variety of data. Time is a crucial factor, machine translation has the ability to save this important time since one does not have to spend time over dictionaries to translate a sentence which will increase productivity.

### B. Related Works

To make machine intelligent natural language processing is used. Language translation is getting improved but there is no significant improvement in English to Bangla machine translations. Researchers proposed many solutions for machine translation. Here for English to Hindi translations, two encoder-decoder neural machine translation architectures are used, which are convolutional sequence to sequence model (ConvS2S) and recurrent sequence to sequence model (RNNS2S) [5]. One is for English to Hindi and another is to do the opposite. In training data, 1492827 sentences are used where 20666365 words for English and 22164816 words for Hindi. The RNNS2S model was trained using the Nematus framework and the ConvS2S model was trained using Fairseq-5, an open-source library developed by Facebook for neural machine translation using Convolution Neural Network (CNN) or RNN networks. Their result showed that ConvS2S performed better on English to Hindi translation which would help to solve our problem. In the corpus-based method using one subject file and one verb file, the translations are solved [4]. Here for each subject, there is a flag corresponded to its verb and the most suitable and meaningful sentences are selected for final translations. The result showed better performance compared to Google Translator.

For another English to Hindi translation feed-forward back-propagation artificial neural network was used [6]. For the implementation, java is used for the main programming language to implement the rules and all the modules apart from the neural network model which have been implemented in Matlab. Here, training data is encoded into numeric form by the Encoder which is also implemented in Java. They have used BLEU [7] to calculate the score of the system. BLEU scores are also applied for testing the training models. Another approach for English to French neural machine translation, RNN encoder-decoder framework methods are implemented [8] where some training procedures and datasets are used to implement for both of those models. After performing the test, the RNN search provided a better result than conventional RNN encode.

*C. Research Objective*

The aim is to design an architecture of English to Bangla Machine Translation system with the recurrent neural network (RNN).

- To synthesize the parsing factors and attention weights of words.
- To design machine translation system architecture with RNN.
- To specify the performance of the machine translation system.

## II. METHODOLOGY

*A. Data Collection*

For training with a machine learning algorithm, datasets are collected. The main dataset of our research is English and Bangla parallel sentences. For each English sentence, we need some co-responding Bangla Sentence to train and test the intelligent system. Dataset is collected from some articles which are manually written in English and Bangla by humans. The maximum lengths of English and Bangla sentences are 7 and 8, respectively.

*B. Sampling*

Dataset consists of 4000 English and Bangla parallel sentences. The dataset is splitting into 80:20 ratio for training and testing.

TABLE I: VOCABULARY DETAILS

| Total English words | 19606 |
|---|---|
| Total Bangla words | 19000 |
| Unique English words | 2839 |
| Unique Bangla words | 3527 |

*C. Preprocessing*

For normalizing the dataset, some text preprocessing steps are done. All the letters of sentences are converted into lowercase and all the punctuations are removed. The characters which do not belong to English and Bangla letters are also dropped out.

*D. Tools*

For model design prototyping, Python and Anaconda Jupyter Notebook are used. The design model of the neural network is developed with Tensorflow-Keras python package distribution.

## III. THE MODEL

*A. Parsing Factor and Tokenization*

Dataset has to be tokenized in the initial state. For each English and Bangla sentences, all the words are tokenized according to the frequency. Tensorflow has a tokenizer library which is used for mapping a word with a corresponding integer number.

TABLE II: TOKENIZED MAPPING WORD

| Sentence | Tokenized Mapping |
|---|---|
| Let me go. | {'let': 5, 'me': 11, 'go': 33, '.': 1} |
| আমাকে যেতে দাও। | {"আমাকে": 7, 'যেতে': 21, 'দাও': 14, '।': 1} |

Then all the words are replaced with a token number and stored in a list for both English and Bangla sentences. The above English sentence of Table II is converted like this-

| 5 | 11 | 33 | 1 |

The corresponding Bangla tokenized sentence looks like-

| 7 | 21 | 14 | 1 |

The length of each sentence is not fixed. To make all the tokenized sentences in the same length, padding is applied. The maximum lengths of English and Bangla sentences are 7 and 8, respectively. So the unfilled blanks are filled with 0 (zeros).

English tokenized sequence:

| 5 | 11 | 33 | 1 | 0 | 0 | 0 | 0 | 0 | 0 |

Bangla tokenized sequence:

| 7 | 21 | 14 | 1 | 0 | 0 | 0 | 0 | 0 | 0 |

The input data is shaped and vectorized, so the dataset is ready for input in our neural network model.

*B. Context Vector with English and Bangla Vocabulary*

With the stored context vector English tokenized sentences would be predicted. To make context vector English and Bangla tokenized sentences are given as input. English and Bangla mapped tokens are made attention weights. Scores are measured for encoder and attention weights. Attention weights represent the attention of the Bangla tokenized sequence over the English tokenized sequence.

$$score = sigmoid(denseLayer + hiddenLayer)$$
$$attentionWeights = softmax(score)$$
$$contextVector = (attentionWeights ? encoderOutput) \quad (1)$$

The context vector is made with Eq. (1), Fig. 1 shows the network for making context vector. The inputs for training sequences are English-Bangla parallel sentences. The embedding layer of RNN normalizes the sequences of token and gives the output as GRU/LSTM layer input. To measure the performance, both GRU and LSTM are implemented and tested. Activating the outputs of dense layers and hidden layers with sigmoid and softmax activation functions, the score for attention weights are measured for performance evaluation. A context vector is made by multiplying encoder outputs and attention weights.

*C. Recurrent Neural Network Model*

The recurrent neural network (RNN) model [9] takes sequential input. The output of a node goes as an input/bias of another node. As words of a sentence have correlational meaning, so the RNN model is used for our study.

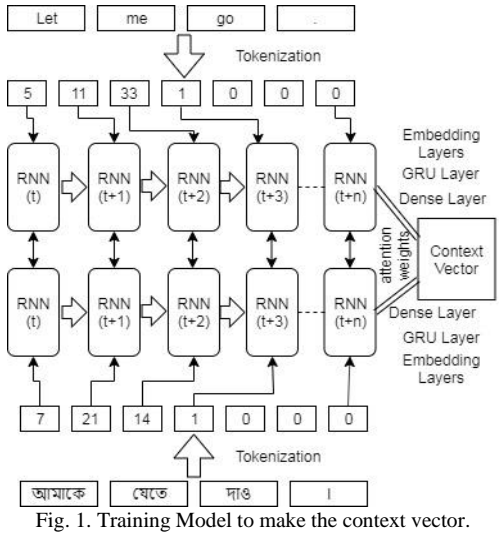

Fig. 1. Training Model to make the context vector.

*1) Encoder*

The encoder is built with combinations of some input embedding layer, GRU layer, and hidden input layers. Embedding layer is used for normalizing the dataset. Gated Recurrent Unit (GRU) [10] is used for the second layer. For testing performance, Long Short Term Memory (LSTM [11] is also used instead of GRU. Different activation functions are used to measure the performance of the model. The encoder output is shaped with batch size 64, sequence length 7, and units 1024.

*2) Decoder*

The first layer of decoder is built with the embedding layer. Then GRU is used as similar to the encoder. An activation function works on GRU layer. After GRU, the dense layer is made with total vocabulary size. Similar to Bahdanau Attention Theory is used, to make the context vector [8].

*3) Attention method*

The attention technique is used for our design. English words focused on which Bangla words are measured with attention weights [8]. The sigmoid activation function is used to normalize inputs with two dense layers and evaluate the score weights. The weights are also normalized with softmax or sigmoid activation function.

*4) Activation function*

To make performance comparison some activation functions are used. The main role of the activation function is to normalize the input sequence.

Tanh Activation Function:

$$F(x) = \frac{Exp(2x) - 1}{Exp(2x) + 1} \quad (2)$$

where *x* is the value of the sequence.

Linear Activation Function:

$$F(x_i) = w_i x_i + b \quad (3)$$

Softmax Activation Function:

$$F(x_i) = \frac{Exp(x_i)}{\sum_{j=1}^{k} Exp(x_j)} \quad (4)$$

Sigmoid Activation Function:

$$F(x) = \frac{1}{1 + Exp(-x)} \quad (5)$$

where $i = 1,...,k$ and $x = (x_1,...,x_k) \in \mathbb{R}^k$ tokenized sequence. The plotting of Softmax and Sigmoid is shown in Fig. 2.

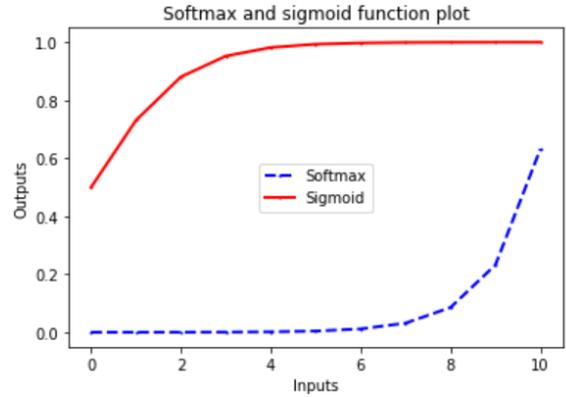

Fig. 2. Sigmoid and Softmax activation function.

All these activation functions are used in the Encoder GRU layer, Decoder GRU layer and attention layer for finding the attention weights.

*5) Loss function*

To train the RNN model, the errors are calculated and improved the model according to errors with a back-propagation loss function. Tensorflow has a sparse categorical cross-entropy function library which is used to calculate the error.

Categorical Crossentropy Function:

$$J(w) = -\frac{1}{N}\sum_{i=1}^{N}[y_i \log(y_i) + (1 + y_i)\log(1 - y_i)] \quad (6)$$

Where *w* is the model parameter or weight of neural network, $y_i$ is the true label and $\hat{y}_i$ is the predicted label of output. The target is to minimize the loss to get a better translation.

*6) Optimization and learning rate*

Dataset preprocessing and normalization are included as the optimization part. The other factor of optimization is the learning rate.

$$\text{LearningRate} = 1e - 3 = 0.001$$

Adam optimization algorithm is used for the learning dataset [12]. Adam is a combination of RMSprop and Stochastic Gradient Descent. The main advantage of using Adam in our study is time optimization during the training dataset.

For the English input sentence- "Let me go", the overall model is shown in Fig. 3. The block of context vector section represents the trained model dataset of mapped English and Bangla tokenized sentences which are defined in the context vector section. Once the input sentence is tokenized and padded, the sequence is ready for RNN input.

In RNN encoder, the first layer is the embedding layer, the second layer is the GRU layer.

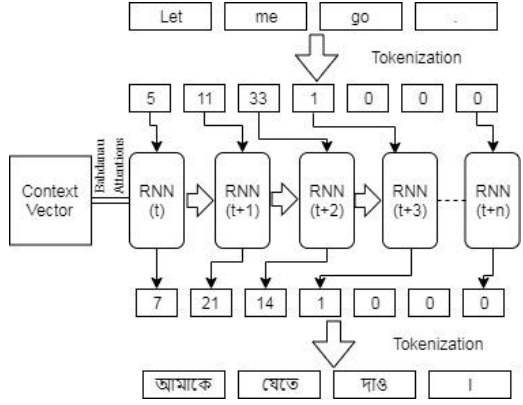

Fig. 3. English to bangla MT model.

The decoder has almost similar to encoder but it has also a dense layer. This layer returns the corresponding token sequence according to the attention score. Then after removing the padding, the tokenizer decodes the sequence into Bangla sentence.

### D. Performance and Accuracy Analysis

Our model has some checkpoints, where the objects of models are saved in local drive, and restore from the previous checkpoint. For every time of training data, it helps to increase the performance. The performance of the model is measured with loss function Eq. (6). 30 epochs are done to evaluate performance. The accuracy of the model depends on the error. Different approaches are performed to minimize the error of the system. Activation functions on the encoder, decoder and attention layers are compared. The performance of GRU and LSTM are also measured here with the best performing activation functions.

*1) Input and output layer*

For the input layer, two activation functions perform best as our experiment. The first layer of encoder is embedding layer, where the tokenized sequences are normalized. After converting all sequences into the embedded format, in the GRU layer tanh Eq. (2) and linear Eq. (3) are tested to measure the performance of the system.

From Fig. 4, the linear activation function for encoder and tanh for decoder GRU layer has stable performance. For crosschecking the error and performances of tanh and linear activation function, four combinations are evaluated.

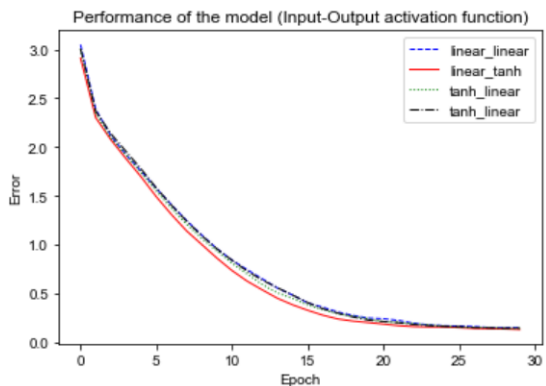

Fig. 4. Loss of the model.

Table III states that, linear activation function for both encoder GRU and decoder GRU increases the error 0.805, for encoder linear and decoder tanh the mean error is 0.740. The other encoder tanh and decoder linear get 0.783 loss.

*2) Attention layer*

The attention layer consists of two activation functions. One is for input and the other one is for normalizing the outputs as attention weights. Sigmoid Eq. (5) and Softmax Eq. (4) are used here. All combinations of sigmoid and softmax are tried to evaluate the best performance.

TABLE III: VALUE LOSS MEAN

| Activation Functions (Encoder-Decoder) | Mean Error (out of 100) | Std. Dev. |
|---|---|---|
| Linear -Linear | 0.805 | 0.787 |
| Linear-Tanh | 0.740 | 0.770 |
| Tanh-Linear | 0.783 | 0.781 |
| Tanh-Tanh | 0.799 | 0.790 |

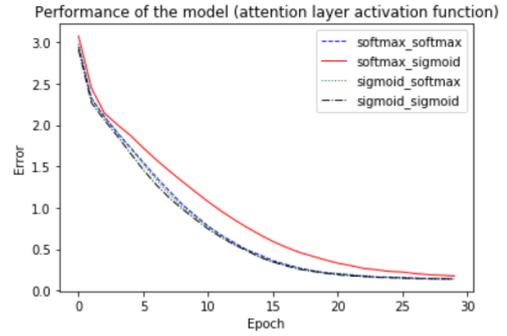

Fig. 5. Loss of model based on attention layer activation function.

Fig. 5 represents, the sigmoid function gives the best performance for the attention layer. Then sigmoid for inputs of attention layer and softmax for output attention layer are efficient.

*3) LSTM and GRU*

After the embedding layer, the GRU layer is implemented. Instead of using GRU, the LSTM layer can be also used. 50 epochs are executed to generate the Fig. 2 and the best performing activation functions are used from the input-output layer and attention layer. But if the epochs have increased, the errors and losses of Gated recurrent unit (GRU) decrease. To get the best performance, some parameters like- Central processing unit (CPU), Random-access memory (RAM) and Graphics processing unit (GPU) and obviously the number of datasets as sample training matters.

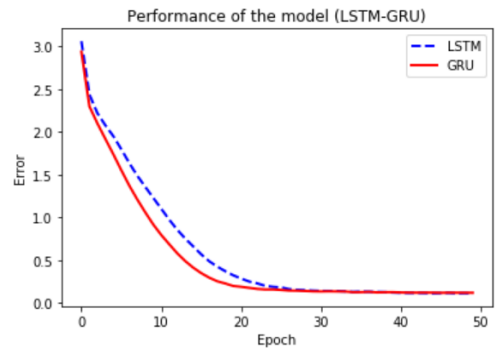

Fig. 6. Performance of the model (LSTM-GRU).

Here in Fig. 6, the performance of GRU looks better than LSTM. That why our final model added GRU layer. The mean error of GRU is 0.508, which is more efficient than LSTM 0.602.

*4) Error minimization per epoch*

100 epochs are done to evaluate the minimization of error for each epoch. Those are divided into two parts, the first 50 and the second 50 epochs. Fig. 7 shows that for the last fifty epochs, the error decreases in satisfaction level.

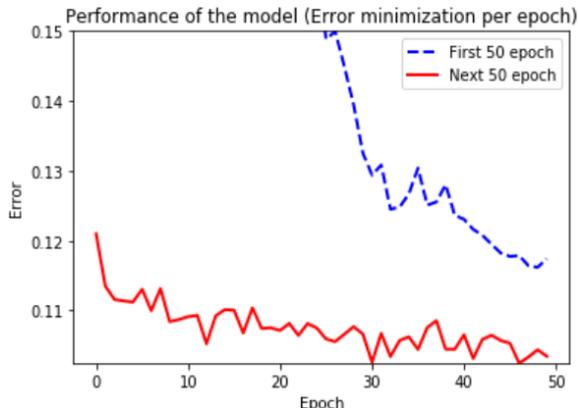

Fig. 7. Performance increase per epoch.

TABLE IV: PERFORMANCE EVALUATION PER EPOCH

| Total (100) epoch | Mean Error | Std. Dev. |
|---|---|---|
| First 50 epoch | 0.506 | 0.680 |
| Next 50 epoch | 0.107 | 0.003 |

From Table IV, for the first 50 epochs, the standard deviation is like 0.680. Approximately the error decreases 0.680 per epoch for the first 50 epochs. For the next 50 epochs, the mean error decreases to 0.17 which is a good performance with 0.003 standard deviations.

## IV. DISCUSSION

The model of our system can evaluate a parallel Bangla sentence based on the input English sentence. This model is generated in a data-driven manner using a deep learning approach. It reforms the words into vector representations by learning to predict translated words integrated with each given word using multiple layers of neural network. After tokenization, the RNN model has embedding layers. Embedding layers are the preliminary layers for both encoder and decoder. To evaluate automatic speech recognition, Benjamin used GRU and LSTM layers performance comparisons and found GRU performed better than LSTM [13]. As our study results state, the performance of the GRU layer is better than the LSTM, so the next layer is the GRU layer. German researcher Bahdanau [8] evaluated the attention with tanh and softmax activation function. For activation function of attention layer sigmoid function is used here for achieving the best performance for Bangla Translation. Two GRU layers of encoder and decoder are activated with linear and tanh activation function as the mean loss for these are minimized. The best accuracy is achieved when in Encoder and Decoder the linear and tanh functions are used respectively and in the attention layer, sigmoid function is used. 100 epochs are done with these configurations and the mean error is minimized in 0.107. Comparing with research [14], our deep learning method has better performance than other machine learning methods. Researcher Arden evaluates the performance of ReLU activation function [15] but according to our research for English to Bangla translation tanh and linear activation works well in the encoder-decoder layer. Our verification shows that the proposed computational technique acquires better performance as a traditional translation. In addition, by considering communicative translation this approach offered the advantage of getting the exact contextual meaning of the given sentence.

As mentioned earlier, getting the actual meaning of a sentence after translation is a complicated task and it depends on the training dataset, vocabulary and CPU processing power. A rich vocabulary can give better performance. Beyond this limitation, our proposed model could be imposed on various applications. Many of the existing solutions focused on the word to word or literal translation without considering the usage of words in a sentence or phrase. The model offers a new possibility to get a well-balanced translation from English to Bangla. Moreover, as the model conquered the well-balanced translation, we believed that it could be adopted to build advance tools or systems that will be able to translate language more accurately by considering the actual meaning of words accordingly.

For example, this model could be used as a conversation system to express emotions and thoughts in Bangla languages without actually knowing the language. This could be a beneficial guide for a lot of individuals. Furthermore, through learning the correct translation, the conversation system will be able to achieve the right meaning of a sentence.

## V. CONCLUSION AND FUTURE WORK

There are lots of benefits of machine translation. It saves time, has the ability to translate many languages, etc. This paper tried to provide a design of machine translation from English to Bangla and implemented them. Our English to Bangla RNN based method provides better results in comparison with other implemented methods found in various research. If the target is fulfilled, it could contribute more to the natural language processing in machine learning algorithms.

After completing the simulation, some lacking are found, which would be solved in the future like- dealing with a large amount of vocabulary can improve the performance, increasing the number of epochs can raise accuracy. Working with these types of issues needs more processing power and memory. The performance can be increased by implementing multiple dense layers with the seq2seq model. Instead of having some issues this study can improve the structure of English to Bangla universal machine translation systems.

CONFLICT OF INTEREST

The authors declare no conflict of interest.

## AUTHOR CONTRIBUTIONS

Shaykh Siddique conducted the research, designed the model, analyzed the data and wrote the section of model and discussion part. Tahmid Ahmed wrote the related works, and conclusion part of the paper. Md. Rifayet Azam Talukder wrote the background section. Md. Mohsin Uddin helped to generate some ideas, critically reviewed the whole paper, fixed the presentation, and grammatical errors of the paper.

## REFERENCES

[1] S. K. Chatterji, "Bengali phonetics," *Bull. Sch. Orient. African Stud*, vol. 2, no. 1, pp. 1–25, 1921.
[2] R. Collobert and J. Weston, "A unified architecture for natural language processing," 2008, pp. 160–167.
[3] K. Muntarina, G. Moazzam, and A. Bhuiyan, "Tense based English to Bangla translation using MT system," *Int. J. Eng. Sci. Invent.*, vol. 2, no. 10, pp. 30–38, 2013.
[4] S. Nahar, M. N. Huda, M. Nur-E-Arefin, and M. M. Rahman, "Evaluation of machine translation approaches to translate English to Bengali," in *Porc. 20th Int. Conf. Comput. Inf. Technol. ICCIT 2017*, pp. 1–5, 2018.
[5] S. Singh, R. Panjwani, A. Kunchukuttan, and P. Bhattacharyya, "Comparing recurrent and convolutional architectures for English-Hindi neural machine translation," in *Proc. 4th Work. Asian Transl.*, no. 2014, pp. 167–170, 2017.
[6] S. Shahnawaz and R. B. Mishra, "A neural network based approach for English to Hindi machine translation," *Int. J. Comput.* vol. 53, no. 18, pp. 50–56, 2012.
[7] K. Papineni, S. Roukos, T. Ward, W. Zhu, and Y. Heights, "IBM research report bleu: A method for automatic evaluation of machine translation," *Science*, vol. 22176, pp. 1–10, 2001.
[8] D. Bahdanau, K. Cho, and Y. Bengio, "Neural machine translation by jointly learning to align and translate," 2014.
[9] Y. K. Tan, X. Xu, and Y. Liu, "Improved recurrent neural networks for session-based recommendations," in *Proc. ACM Int. Conf. Proceeding Ser.*, vol. 15-September, Sep. 2016, pp. 17–22.
[10] R. Dey and F. M. Salemt, "Gate-variants of gated recurrent unit (GRU) neural networks," in *Proc. Midwest Symposium on Circuits and Systems*, 2017, pp. 1597–1600.
[11] M. Sundermeyer, R. Schlüter, and H. Ney, "LSTM neural networks for language modeling," 2012.
[12] D. P. Kingma and J. Lei Ba, "Adam: A method For stochastic optimization," 2014.
[13] S. Khandelwal, B. Lecouteux, and L. Besacier, "Comparing gru and lstm for automatic speech recognition," 2017.
[14] Y. Wu *et al.*, "Google's neural machine translation system: Bridging the gap between human and machine translation," Sep. 2016.
[15] M. Karpinski, V. Khoma, V. Dudvkevych, Y. Khoma, and D. Sabodashko, "Autoencoder neural networks for outlier correction in ECG- Based biometric identification," in *Proc. 2018 IEEE 4th International Symposium on Wireless Systems within the International Conferences on Intelligent Data Acquisition and Advanced Computing Systems,* 2018, pp. 210–215.

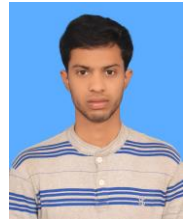

**Shaykh Siddique** was born in Mymensingh, Bangladesh, in 1997. He received higher secondary certificate (HSC) from Govt. Shahid Smriti College, Mymensingh, Bangladesh in 2015. Since 2016, he is continuing his bachelor of science (BSc) degree in Department of Computer Science and Engineering, East West University, Dhaka, Bangladesh. He also participated in the 2017 ACM-ICPC Bangladesh, CUET NCPC Provincial Programming Contest. His research interests include algorithms, machine learning, distributed computing, cybersecurity, and software modeling.

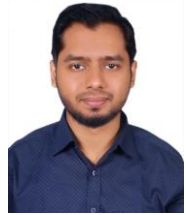

**Tahmid Ahmed** was born in Dhaka, Bangladesh in 1998. He received in HSC and the SSC degrees from National Ideal College, Dhaka, Bangladesh in 2015 and 2013, respectively. In 2016, he admitted to the Department of Computer Science and Engineering at East West University, Dhaka, Bangladesh as an undergraduate student. He is interested in machine learning, app development, computer networks, and pattern recognition.

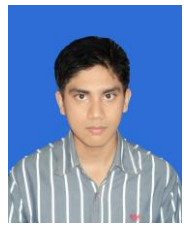

**Md. Rifayet Azam Talukder** was born in Rampur, Bangladesh in 1997. He received his HSC and SSC degrees in 2014 and 2012, respectively. In 2016, he admitted at East West University, Dhaka, Bangladesh as an undergraduate student focused on Computer Science and Engineering. He is also an undergraduate teaching assistant at East West University under the Computer Science and Engineering department. He is also interested in the machine learning field. Currently he is working on deep neural network, and RNN.

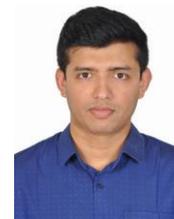

**Md. Mohsin Uddin** is currently a lecturer of the Department of Computer Science and Engineering at East West University, Dhaka. He joined East West University in April 2018. He has obtained BSc. Engg. degree in computer science and engineering from Bangladesh University of Engineering and Technology (BUET), Dhaka. He has obtained his MS degree in computer science from the University of Lethbridge, Alberta, Canada. He worked as a software engineer in various multinational software companies in Canada and Bangladesh. He has 6+ years of professional experience in big data, NLP, machine learning systems design and software development. He has several research publications in renowned international conferences. His research areas are NLP, machine learning, and big data.